\documentclass[sigconf]{acmart}
\usepackage{algorithm}
\usepackage{algorithmic}
\usepackage{diagbox}
\usepackage{breqn}
\usepackage{graphicx}
\usepackage{fancyhdr}
\AtBeginDocument{%
  }

\copyrightyear{2026}
\acmYear{2026}
\acmConference[The Web Conference]{Enhancing Web Recommendation Roubustness via Scenario Knowledge Transfer for Cold Users}{April 13-17, 2026}{Dubai, United Arab Emirates}
\fancypagestyle{firstpage}{
  \fancyhf{}
  \fancyhead[L]{\footnotesize Published as a conference paper at WWW 2026}
}
\begin{document}

% \title{Enhancing Web Recommendation Roubustness via Scenario Knowledge Transfer for Cold Users}
\title{Trinity: A Scenario-Aware Recommendation Framework for Large-Scale Cold-Start Users}
\author{Wenhao Zheng, Wang Lu, Fangshuang Tang, Yiyang Lu, Jun Yang, Pengcheng Xiong, Yulan Yan}
\email{{wenhaozheng, wanlu, fredtang, yiyanglu, yajun, peterxiong, yulanyan}@mircosoft.com}

\affiliation{%
  \institution{Mircosoft Corp}
  \city{Redmond}
  \state{WA}
  \country{USA}
}
%\thanks{Corresponding author is Pengcheng Xiong}

% \affiliation{%

% \affiliation{%

\renewcommand{\shortauthors}{Wenhao et al.}
%\thanks{published as a workshop paper at WWW 2026}
\begin{abstract}
  % In products equipped with recommendation systems, there is often a need to introduce new scenarios to attract new user groups. In the initial stage of a new scenario, it is typically dominated by new users. This introduces two major challenges: first, the behavioral data of new users is sparse; second, the data distribution in the new scenario undergoes shifts. Previous works often rely solely on model architecture to address only part of these issues. We argue that the recommendation problem involving new users in new scenarios should be tackled by integrating feature engineering, model architecture, and training paradigms. To this end, we propose a novel framework Trinity, which means aboves three components are inseparable and work synergistically. It could extract valuable information from existing scenarios while ensuring the prediction effectiveness and accuracy in the new scenario. Both offline and online experiments demonstrate that our framework achieves remarkably significant improvements in addressing the new user plus new scenario problem.
Early-stage users in a new scenario intensify cold-start challenges, yet prior works often address only parts of the problem through model architecture. Launching a new user experience to replace an established product involves sparse behavioral signals, low-engagement cohorts, and unstable model performance. We argue that effective recommendations require the synergistic integration of feature engineering, model architecture, and stable model updating.
We propose Trinity, a framework embodying this principle. Trinity extracts valuable information from existing scenarios while ensuring predictive effectiveness and accuracy in the new scenario. In this paper, we showcase Trinity applied to a billion-user Microsoft product transition. Both offline and online experiments demonstrate that our framework achieves substantial improvements in addressing the combined challenge of new users in new scenarios.
\end{abstract}

%\end{CCSXML}




\maketitle
\thispagestyle{firstpage}
\section{Introduction}
% With the advancement of deep machine learning technologies, an increasing number of recommendation platforms have extensively adopted corresponding techniques, enabling users to discover interested items while generating substantial commercial value for platforms. To capture the diverse needs of different user groups, platforms often establish multiple scenarios. Each of them has distinct page styles and partially overlapping recommendation items. This gives rise to two major challenges: first, how to migrate user behaviors from classic scenarios to new ones; and second, how to ensure the stability and accuracy of recommendation performance for cold-start users in these new scenarios.
In the technology industry, it is common for products to undergo continuous evolution. As user behaviors, market demands, and technological capabilities change, products must adapt to remain relevant. This often involves updating core features, redesigning user experiences, integrating new technologies, or even pivoting to entirely new use cases. Microsoft MSN provides a clear example of this phenomenon. Since its launch in 1995 as an online service and ISP, MSN has continuously transformed—shifting from a dial-up service to a web portal, rebranding under Windows Live, evolving its search capabilities into Bing, and most recently incorporating AI-driven recommendations and Copilot Discovery. These changes reflect MSN’s ongoing adaptation to user needs and technological trends, illustrating the broader challenge of managing product evolution in a dynamic digital landscape.

The product evolution often brings technical challenges, particularly in adapting recommendation systems to new scenarios and user behaviors. With the advancement of deep learning technologies, recommendation platforms increasingly adopt these techniques to help users discover items of interest while generating substantial commercial value. To cater to the diverse needs of different user groups, platforms often establish multiple scenarios, each with distinct page layouts and partially overlapping content. This introduces two key challenges: (1) how to transfer user behaviors from established scenarios to new ones, and (2) how to ensure stable and accurate recommendations for cold-start users in these new scenarios.

\begin{figure}[t]
\centering
\includegraphics[width=1\linewidth]{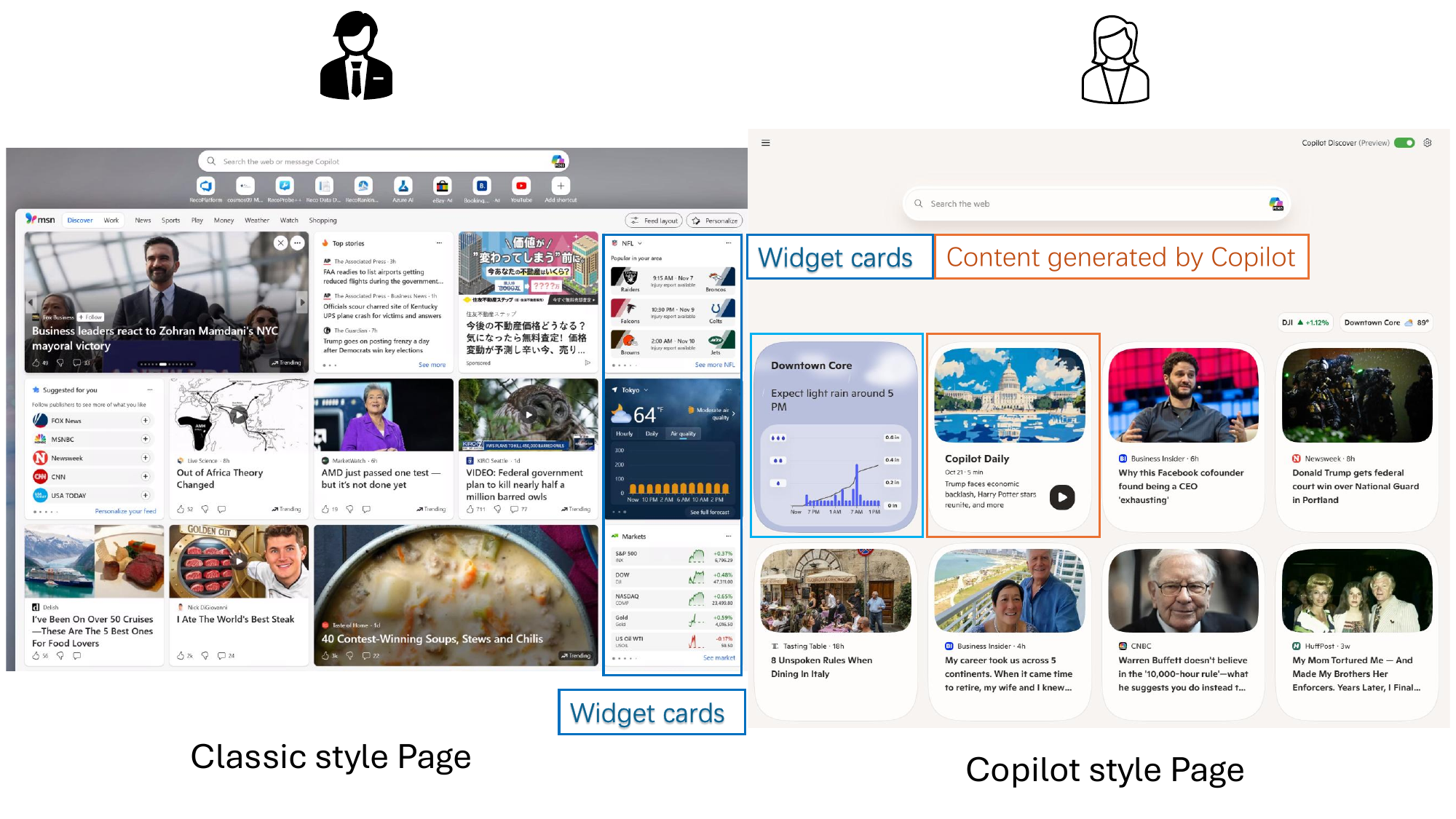}
% \caption{The layouts and contents vary across different style pages, which is crucial for user click-through rates.}
\caption{Microsoft MSN product migration from classic style (left) to copilot style (right).}
\label{fig:example}
\end{figure}

Figure~\ref{fig:example} illustrates two styles of Microsoft MSN products: the left side shows the classic style with fixed news and widgets, while the right side represents the Copilot style, which selectively displays Copilot-related content and omits some widgets. Although the pages share overlapping content, each scenario has unique elements, posing challenges for recommendation algorithms. Most training data comes from the classic scenario, while the Copilot-style scenario is sparse and dominated by cold-start users. Additionally, context changes can shift click-through rates for identical content, requiring careful calibration to ensure stable and accurate recommendations across scenarios.

Existing solutions provide limited support for the challenges described above. Current multi-scenario approaches, such as MMoE~\cite{ma2018modeling} and PLE~\cite{tang2020progressive}, employ task-specific gating mechanisms to handle multi-scenario and multi-objective learning. However, when faced with severe sample imbalance across scenarios, these gates often fail to fit effectively, resulting in degraded performance in new scenarios. PePNet~\cite{chang2023pepnet} incorporates scenario-feature interactions but neglects scenario signals within the dense tower, causing predictions in new scenarios to be biased toward dominant old scenario patterns. Moreover, these methods typically follow conventional daily continuous model update schedule, leading to highly unstable model performance in production environments.

In this paper, we propose a novel framework that tackles these issues from three complementary perspectives: features, model architecture, and model update.

First, at the feature level, traditional methods focus solely on features related to the current target item. We posit that correlations exist across different content, and cold-start users often exhibit behaviors specific to particular content. Capturing features across all content is therefore essential.

Second, in terms of model architecture, it is necessary to strengthen scenario and card type signals, enabling them to effectively filter and enhance feature representations. Additionally, because click biases vary significantly across scenarios, calibration based on scenario information is required to ensure that the overall Click Over Predicted Click (COPC) remains well-aligned.

Finally, conventional recommendation algorithms rely on daily or hourly updates. In new scenarios dominated by cold-start users, this can lead to model jitter, trapping the model in local optima and causing gradual online performance degradation. We propose a dynamic model update approach to maintain robust model training despite fluctuations in user behavior.

To the best of our knowledge, this is the first paper to document industry practices for billion-scale new product migration. We hope our experience can help open new research directions and inform future engineering advancements. The remainder of this paper is organized as follows. In Section~\ref{sec:framework}, we present the main framework. Section~\ref{sec:exp} reports substantial improvements demonstrated through both offline and online experiments.
Finally, we conclude in Section~\ref{sec:conclude}.
%We discuss related work in Section~\ref{sec:relate}, and finally, we conclude in Section~\ref{sec:conclude}.

\begin{figure*}[htpb]
\centering
\includegraphics[width=0.7\linewidth]{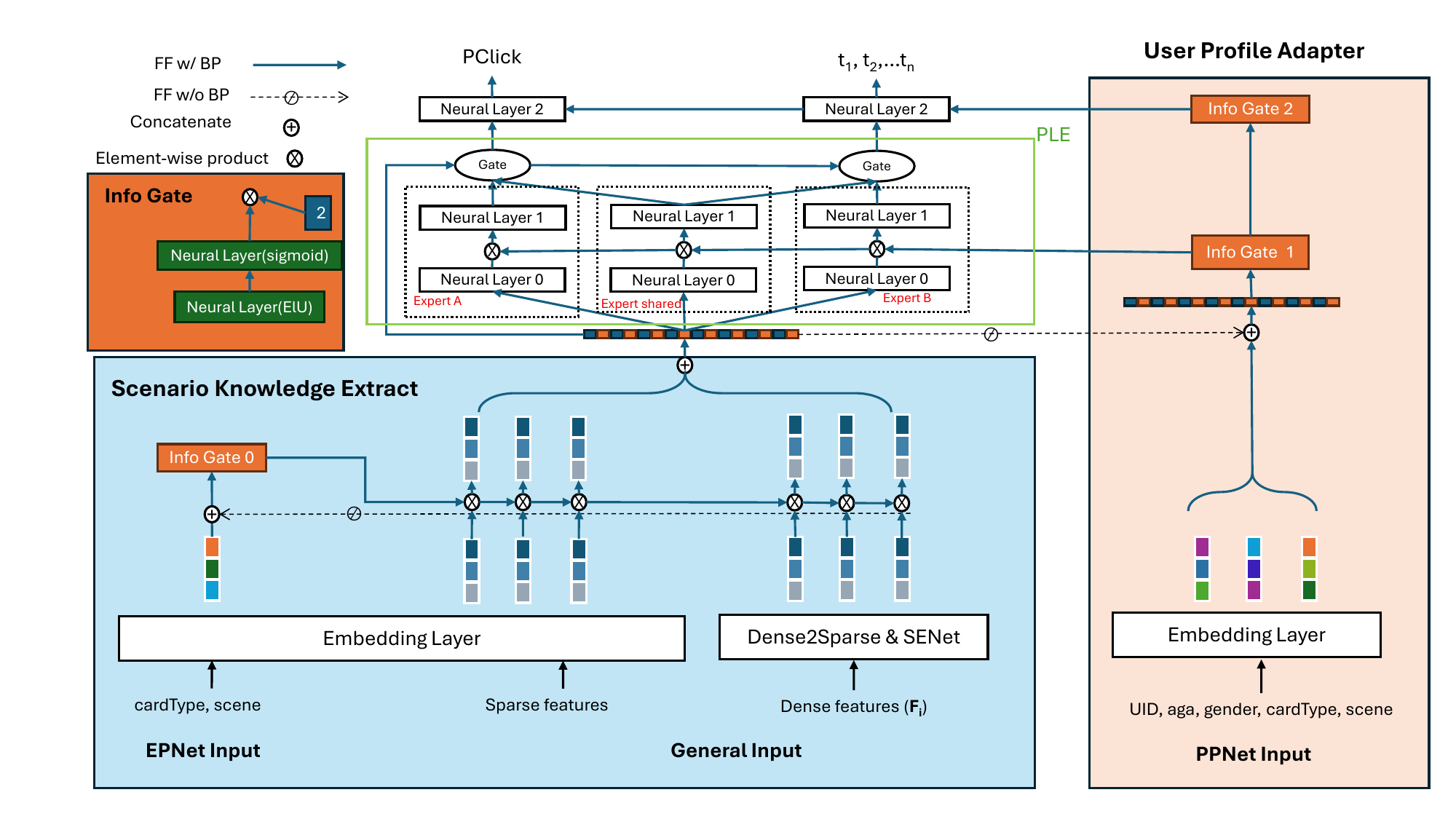}
\caption{The model structure. This model is a multi-task model, where $t_1, t_2, \dots, t_n$ represent multi-classified durations. Since duration is similar to clicks and due to space limitations, duration-related content will not be further elaborated.}
\label{fig:framework}
\end{figure*}

\section{Main Framework}\label{sec:framework}
% To effectively address the cold-start user recommendation problem in the Edge homepage Copilot style, our framework consists of three components. First, in the feature engineering part, we incorporate all user × cardtype features to construct a relatively complete user interest profile. Second, inspired by PePNet, we enhance feature selection via scene signals and calibrate the predict values for new scenes. Third, we adopt a more robust training approach to counteract the drift of data distribution in new scenes.
Our framework adheres to standard engineering principles in large-scale recommendation systems and is organized around three key components: feature design, model architecture, and model updating.

\subsection{Feature Engineering}\label{sec:feature}
In conventional recommendation systems, the large scale of candidate items often leads to retaining only user behavior features directly associated with the interactive (target) item. However, in cold-start scenarios—where items are newly introduced and users are relatively inactive—such target-specific behavioral signals are extremely sparse, making it difficult for the model to learn reliable preference patterns. For example, in the Copilot-style scenario, Copilot-generated content cards appear exclusively within the Copilot layout and do not exist in classic-style browsing contexts, resulting in no historical interaction signals for most users. This motivates incorporating user interaction features across all scenarios, rather than restricting to features aligned only with the target item.

Existing approaches commonly model user preferences via sequential behavior features. However, sequential features present two inherent limitations under cold-start conditions. First, inactive users have short and sparse interaction histories, providing insufficient signal to form effective behavioral associations with candidate items. Second, active users in mature (classic) scenarios possess substantially longer behavior sequences, which dominate model optimization, causing the representation to overfit to active-user patterns and weakening generalization to inactive or new users. As a result, reliance on sequential features leads to degraded performance in cold-start environments.

To address these challenges, we construct a statistical behavior feature tensor $\mathbf{F}_i \in \mathbb{R}^{ |\mathcal{T}| \times |\mathcal{S}| \times |\mathcal{C}| \times |\mathcal{A}|}$ for each user $U_i$, where $\mathcal{T} = \{1\mathrm{h},\;1\mathrm{d},\;7\mathrm{d},\;30\mathrm{d}\}$ represents the time dimension, $\mathcal{S} = \{\text{classic},\;\text{copilot},\;\text{all}\}$ denotes the scenario dimension, $\mathcal{C} = \{\text{weather}, \allowbreak \;\text{finance},\;\text{news},\;\text{video},\;\text{copilot-content}\}$ denotes the card type and $\mathcal{A} = \{\text{view},\;\text{click}\}$ denotes user action type.
This formulation ensures that new and existing users share a unified and fixed feature space, while capturing cross-scenario behavioral signals necessary for robust preference modeling in cold-start settings.

\subsection{Model Architecture}
To fully exploit the proposed statistical behavior representation, we design a model architecture that amplifies signals from emerging scenarios while robustly characterizing users under cold-start conditions. As shown in Figure~\ref{fig:framework}, the architecture comprises two main components. The \textit{Scenario Knowledge Extractor} stabilizes dense feature representations and adaptively emphasizes scenario-specific signals. 
The \textit{User Profile Adapter} reinforces card-type and scenario signals once more, ensuring that final predictions are more calibrated and accurate.
% In addition, card-type and scenario embeddings are included to gate all feature signals, allowing precise modeling of card-type-specific variances between scenarios. Subsequently, the 

\subsubsection{Scenario Knowledge Extractor}
User interaction signals are inherently discrete and highly variable, and outliers may adversely affect model convergence. To obtain stable representations, we convert dense interaction statistics into sparse embeddings through a learned dense-to-sparse transformation. Unlike conventional methods that rely solely on equal-width or equal-frequency binning, we first apply Batch Normalization to the dense features before equal-frequency binning. This allows the bucket boundaries to adapt to the underlying data distribution, producing more semantically consistent sparse features.

The feature tensor $\mathbf{F}_i$ constructed in Section~\ref{sec:feature} contains thousands of dense features arising from Cartesian product interactions. To reduce computational cost, we employ a Squeeze-and-Excitation Network (SENet)~\cite{hu2018squeeze} to model inter-feature dependencies and compress the feature representation to one-third of its original dimensionality, improving both training efficiency and inference latency.

Since most behavioral data originate from the classic-style scenario, we introduce card-type and scenario embeddings to perform gated feature modulation. The gating mechanism outputs values in [0, 1] after a sigmoid activation, which would normally center feature strengths near 0.5. To avoid uniformly weakening the signal, we rescale the gating output by a factor of 2, shifting the distribution to center around 1. 
% User actions on items are typically discrete and unstable, with certain outliers potentially hindering model convergence. To mitigate this, we convert discrete values into sparse representations via hash tables to obtain stable embeddings, which is critical for cold-start user recommendations. Unlike conventional dense-to-sparse methods that rely on equal-width or equal-frequency binning, we integrate their strengths by applying BatchNorm to dense features prior to equal-frequency binning, enabling bucket boundaries to adaptively align with the data distribution. Given that Cartesian product-based feature construction ($\mathbf{F}_i$) yields thousands of dense features, we incorporate SENet to compress feature vectors to one-third of their original dimensionality, thereby enhancing training and inference efficiency.

% Since nearly all features are dominated by the classic style scenario, we need to use cardtype and scenario features for filtering to achieve the goal of strengthening the new scenario. After passing through the sigmoid layer of Gate NU, a value between 0 and 1 is obtained. Due to the properties of sigmoid, most values are distributed around 0.5, which weakens all features. Therefore, we multiply by 2 here to make most values distributed around 1, where values greater than 1 indicate strengthening and less than 1 indicate weakening.

\subsubsection{User Profile Adapter}
Even after feature-level gating, classic-style data may still dominate downstream modules such as PLE (need citation) and neural interaction layers, leading to biased outputs (e.g., pClick) in Copilot-style scenarios. Such bias can propagate to downstream re-ranking and traffic shaping modules, negatively affecting overall system performance. To mitigate this, we introduce a \textit{User Profile Adapter}, inspired by PPNet(need citation), which aggregates user profile features, card-type embeddings, and scenario embeddings to recalibrate the outputs of the shared representation layers.

This module ensures that all samples (both old and new scenarios) contribute to parameter learning while aligning predicted scores across scenarios for the same user–item pair. As a result, the model maintains consistent behavioral semantics across scenarios, improving cold-start robustness and preventing performance drift when new scenarios are introduced.

% Even with feature filtering, the classic style data still dominates in PLE and Neural Layer 2, causing the cardtype scoring in the copilot style scenario to bias towards the performance of the classic style. If the prediction drifts, it will affect the subsequent rerank module and traffic shaping module of the recommendation system. Therefore, we aim to avoid this phenomenon. Drawing inspiration from the PPNet approach, we design the User Profile Adapter module, which aggregates user information, cardType, and scenario features to adjust the outputs of PLE and Neural Layer 2.

% Thus, for the parameters of the entire model, all samples participate in the update. Moreover, it ensures that, under the same user and item conditions, the predicted values across different scenarios closely align with the true average values of the respective scenarios.

\subsection{Stability-Aware Model Updating}
% New users in the new scenario exhibit another characteristic: extremely intense behavioral preferences. This results in not all samples being effective; fitting some samples to the model can instead trap it in local optima, leading to continuous degradation of the production model's performance. We innovatively propose the following training update strategy, which effectively prevents abnormal model updates while ensuring a certain level of timeliness.

Users in the new scenario often exhibit highly volatile and rapidly shifting behavioral patterns, which results in large fluctuations in training signals. Consequently, not all daily training data contribute positively to model optimization; in some cases, updates driven by noisy or unstable samples may push the model toward suboptimal local minima and lead to online performance degradation. To mitigate this risk while preserving the timeliness of model updates, we adopt a stability-aware update mechanism that selectively incorporates new checkpoints based on both global ranking performance and scenario-specific calibration.

Specifically, after training a candidate model on the most recent data, we compare it with the current deployed checkpoint using overall Area Under the Receiver Operating Characteristic Curve (AUC) and the Click Over Predicted Click (COPC) metric, which measures the consistency of scenario-level performance. The new checkpoint is only accepted if it improves AUC and does not increase COPC deviation beyond a small tolerance threshold $\delta$ that is optimized by our training data. Otherwise, the previous checkpoint is retained.

% \begin{algorithm}
% \caption{Daily Model Checkpoint Update Strategy}
% \label{alg:update}
% \begin{algorithmic}[1]
% \REQUIRE Initial model checkpoint $M_0$
% \STATE $M_{\text{current}} \gets M_0$
% \FOR{$t = 1, 2, \dots, T-1$}
%     \STATE $AUC_{\text{old}}, COPC_{\text{old}} \gets \text{Evaluate}(M_{\text{current}}, \text{Data}_{t+1})$ \quad \textcolor{gray}{\textit{// Evaluate current checkpoint on day T+1 data}}
    
%     \STATE $M_{\text{new}} \gets \text{Train}(\text{Data}_t)$ \quad \textcolor{gray}{\textit{// Train a new model using day T data}}
    
%     \STATE $AUC_{\text{new}}, COPC_{\text{new}} \gets \text{Evaluate}(M_{\text{new}}, \text{Data}_{t+1})$ \quad \textcolor{gray}{\textit{// Evaluate the new model on day T+1 data}}
    
%     \IF{$AUC_{\text{new}} > AUC_{\text{old}} \quad \textbf{and} \quad  | 1 - COPC_{\text{new}} | <= | 1 - COPC_{\text{old}} | + \delta $}
%         \STATE $M_{\text{current}} \gets M_{\text{new}}$
%         \STATE $\text{SaveCheckpoint}(M_{\text{current}})$
%     \ELSE
%         \STATE $\text{SaveCheckpoint}(M_{\text{current}})$ \quad \textcolor{gray}{\textit{// Keep previous checkpoint}}
%     \ENDIF
% \ENDFOR
% \end{algorithmic}
% \end{algorithm}
\begin{algorithm}
\caption{Daily Model Checkpoint Update Strategy}
\label{alg:update}
\begin{algorithmic}[1]
\REQUIRE Initial checkpoint $M_0$
\STATE $M_{\text{current}} \gets M_0$
\FOR{$t = 1, 2, \dots, T-1$}
    \STATE $(AUC_{\text{old}}, COPC_{\text{old}}) \gets \text{Evaluate}(M_{\text{current}}, \text{Data}_{t})$
    \STATE $M_{\text{new}} \gets \text{Train}(\text{Data}_{t})$
    \STATE $(AUC_{\text{new}}, COPC_{\text{new}}) \gets \text{Evaluate}(M_{\text{new}}, \text{Data}_{t})$
    \IF{$AUC_{\text{new}} > AUC_{\text{old}} \;\textbf{and}\; |1 - COPC_{\text{new}}| \le |1 - COPC_{\text{old}}| + \delta$}
        \STATE $M_{\text{current}} \gets M_{\text{new}}$
    \ENDIF
    \STATE $\text{SaveCheckpoint}(M_{\text{current}})$
\ENDFOR
\end{algorithmic}
\end{algorithm}
This stability-aware updating mechanism allows the model to evolve continuously with newly collected data while preventing performance collapse caused by unstable user behavior fluctuations in emerging scenarios.

\section{Experiment}\label{sec:exp}
In this section, we conduct comprehensive offline and online evaluations on large-scale multi-scenario recommendation systems to assess the effectiveness of the proposed framework. Our primary motivation originates from the Edge homepage scenario, where new layout styles and interaction patterns pose challenging cold-start and cross‑scenario transfer issues. Therefore, the experiments in this section are designed and evaluated primarily around the Edge homepage scenario to reflect real production constraints and validate the practical effectiveness of the framework.

\subsection{Experimental Settings}
In this part, we describe the datasets, baseline methods, hyperparameters, and evaluation metrics used in our experiments.
\subsubsection{Datasets}
\begin{table}[]
\caption{Dataset description.}
\resizebox{0.7\columnwidth}{!}{
      \begin{tabular}{@{}ccccc@{}}
      \toprule
      Dataset      & \# User &  \# Train &   \# Test & pClick \\ \midrule
      Classic style & 280M  & 1.2B  & 0.6B    & 1\%   \\
      Ruby style   & 14M  & 185M  & 95M       & 0.1\%   \\ \bottomrule
\end{tabular}
  }
  \label{data_description}

\end{table}
The dataset is extracted from over 1 billion monthly active user in Microsoft MSN ecosystem. Data from July 14–28 is used for training, and data from August 1–7 is used for testing. To align with real-world deployment practice, model evaluation at each day allows training with all data prior to that day (e.g., AUC on August 5 is computed using data up to August 4). The final performance is reported as the average across the 7 test days.

% The dataset is derived from the Edge homepage scenario, utilizing data from July 14 to 28 for training and from August 1 to 7 for testing. To align with real production applications, evaluation allows the use of prior days for training; for instance, when evaluating the AUC for data on the 5th, data before August 5th can be employed as the training set. The average value over the 7 testing days is taken as the final test result.

\subsubsection{Baselines and variants} 
We compare our framework \textbf{$Trinity$} with the following baselines:  
\textbf{PLE} models multi-task learning via multiple experts.  
\textbf{PePNet} strengthens scenario representation through personalized user embeddings.
We also conduct ablation studies using the following variants of our framework:
\textbf{$Trinity_{small}$} retains only behavior features relevant to the current target item, rather than preserving all behavior features.  
\textbf{$Trinity_{w/o~check}$} removes AUC and COPC validation checks and performs continuous daily updates.  
\textbf{$Trinity_{w~PLE}$} replaces the model backbone with PLE.

\subsubsection{Hyper-Parameter}
We use ELU~\cite{clevert2015fast} as the activation function for all models. The batch size is set to 1024. We adopt the Adam optimizer~\cite{kingma2014adam} with $\beta_1=0.9$, $\beta_2=0.999$, and $\epsilon=1\times10^{-6}$, and we set the learning rate to 0.0001.

\subsubsection{Evaluation Metrics}
To evaluate performance across scenarios, we use AUC and COPC. COPC is defined as $COPC = \frac{realctr}{pctr}$.
A COPC value closer to 1 indicates better calibration between predicted CTR and actual CTR.

\begin{table}[h]
\centering
\caption{ Model performance on the multiple scenarios}
\label{performance compare}
\resizebox{0.7\columnwidth}{!}{
\begin{tabular}{lcccc}
\toprule
Dataset & \multicolumn{2}{c}{Classic style} & \multicolumn{2}{c}{Copilot style} \\
\midrule
\diagbox{\small Model}{\small Metric} & AUC & COPC & AUC & COPC \\
\midrule
PLE & 0.838 & 1.12 & 0.564 & 0.13 \\
PePNet & 0.845 & 1.05 & 0.584 & 0.12 \\
$Trinity_{small}$ & 0.841 & 1.23 & 0.701 & 0.82 \\
$Trinity_{w/o~check}$ & 0.852 & 1.18 & 0.543 & 0.12 \\
$Trinity_{w~PLE}$ & 0.861 & 1.04 & 0.616 & 1.81 \\
$Trinity$ & \textbf{0.869} & 1.02 & \textbf{0.726} & 0.95 \\
\bottomrule
\end{tabular}
}
\end{table}
\subsection{Performance Evaluation}
As shown in Table~\ref{performance compare}, 
the proposed framework consistently outperforms all baseline methods. The ablation results further confirm that each component of our design contributes significantly to performance, with the most notable gains observed in the Copilot-style scenario where cold-start users dominate.

In the Classic style scenario of the MSN product, where abundant historical data is available, all methods achieve AUC values above 0.8 and COPC values close to 1. In this setting, \textbf{Trinity} provides only marginal improvements, which is expected when high-quality and stable training data are available. In contrast, in the Copilot-style scenario characterized by sparse user behavior and shifting distributions, both baselines perform poorly, showing AUC values close to random guessing (around 0.5) and severe overestimation of predicted CTR (low COPC). Our \textbf{Trinity} framework substantially alleviates these issues, achieving an AUC of approximately 0.7 and a COPC close to 1.

The \textbf{$Trinity_{w~PLE}$} variant underestimates predicted CTR, resulting in a COPC significantly greater than 1. This is likely because PLE overemphasizes shared representations learned from the Classic scenario, causing insufficient adaptation to sparse cold-start distributions. The results for \textbf{$Trinity_{small}$} indicate that preserving the full set of user behavior features, rather than only those aligned with the target item, yields further improvements by capturing cross-content user preference patterns.

Finally, \textbf{$Trinity_{w/o~check}$} achieves performance comparable to the weaker baselines. Removing AUC and COPC validation checks leads to unstable daily updates and model drift, demonstrating that robust update strategies are essential. This confirms that incorporating stability-aware model updating is critical to achieving reliable performance in cold-start and evolving product scenarios.
% the framework we propose outperforms all baseline methods, and each of our components is effective, achieving a particularly significant improvement in the Copilot style scenario where cold-start users dominate.

% Almost all methods achieve an AUC above 0.8 and a COPC close to 1 on the classic style, indicating that under sufficient data training, our framework can achieve certain improvements, but the magnitude of improvement is not significant. In contrast, in the cold-start scenario, previous works perform extremely poorly, with predictions severely drifted due to the influence of classic style data. It is noteworthy that without AUC \& COPC checks, the model's training becomes unstable, resulting in very poor performance in the Copilot style.
\subsection{Online A/B Testing}
We conducted large-scale online A/B testing on Microsoft MSN products from 2025-08-19 to 2025-08-23 with real users. The proposed \textbf{Trinity} framework delivered substantial improvements over the existing production system, achieving a \textbf{5.61\%} increase in Time Spent and a \textbf{3.04\%} increase in iDAU\footnote{iDAU refers to interaction DAU; a user must perform at least one meaningful interaction within the MSN ecosystem to be counted.}. This marks the largest performance improvement observed among Copilot-style product iterations in the past six months.

Our Trinity introduces only approximately 10 ms of additional inference latency, which is negligible compared to the overall end‑to‑end pipeline latency of around 300 ms. Moreover, since our training and inference leverage more card‑type interaction features, the storage footprint of the feature set increases by roughly 20\%.

\section{Conclusion}\label{sec:conclude}
In this work, we address the challenges posed by new users and new scenarios by introducing \textbf{Trinity}, a unified framework that integrates feature engineering, model architecture, and training strategy. The proposed framework effectively mitigates the cold-start problem and reduces performance degradation caused by scenario shifts. Both offline evaluations and large-scale online A/B testing demonstrate that each component of the framework is essential and that their combination yields significant performance gains. 
% In future work, we plan to extend the framework to additional product scenarios and further enhance end-to-end recommendation performance by exploiting the behavioral characteristics of cold-start users.

% \begin{acks}
% \end{acks}

\bibliographystyle{ACM-Reference-Format}
\bibliography{sample-base}


\end{document}